\documentclass[runningheads]{llncs}
\usepackage[]{todonotes}
\usepackage[linesnumbered,lined,ruled,commentsnumbered]{algorithm2e}
\usepackage{amssymb}
\usepackage{float}
\usepackage{enumitem}

\setlist[itemize]{noitemsep}
\setlist[enumerate]{noitemsep}

\usepackage[T1]{fontenc}
\usepackage{graphicx}
\usepackage{hyperref}
\begin{document}
    \title{Hybrid Genetic Search for Dynamic Vehicle Routing with Time Windows}

    \author{Mohammed Ghannam\inst{1,2}\orcidID{0000-0001-9422-7916} \and \\
    Ambros Gleixner\inst{1,2}\orcidID{0000-0003-0391-5903}}

    \institute{HTW Berlin, Treskowallee 8, 10318 Berlin, Germany \\
    \and
    Zuse Institute Berlin, Takustr. 7, 14195 Berlin, Germany  \\
    \email{\{ghannam, gleixner\}@htw-berlin.de}
    }
    \maketitle              

    \begin{abstract}
        The dynamic vehicle routing problem with time windows (DVRPTW) is a generalization of the classical VRPTW to an online setting,
        where customer data arrives in batches and real-time routing solutions are required.
        In this paper we adapt the Hybrid Genetic Search (HGS) algorithm, a successful heuristic for VRPTW,
        to the dynamic variant.
        We discuss the affected components of the HGS algorithm including giant-tour representation, cost computation,
        initial population, crossover, and local search.
        Our approach modifies these components for DVRPTW, attempting to balance solution quality and constraints on future customer arrivals.
        To this end, we devise methods for comparing different-sized solutions, normalizing costs, and accounting for
        future epochs that do not require any prior training.
        Despite this limitation, computational results on data from the EURO meets NeurIPS Vehicle Routing Competition 2022
        demonstrate significantly improved solution quality over the best-performing baseline algorithm.


    \end{abstract}

    \section{Introduction}

    Vehicle routing has attracted a lot of attention in the past decades due to its prevalence in many
    branches of industry.
    One of the most prominent variants is the \emph{Capacitated Vehicle Routing problem with Time
    Windows} (VRPTW).
    The problem is defined on a directed graph $G = (V,A)$ where the set of nodes $V$ represents the customers
    and the set of arcs $A$ represents possible connections between nodes.
    Throughout the paper, nodes and customers are used interchangeably.
    Arc weights $c_a > 0$ are defined for all $a \in A$ to represent the cost for driving along the arc.
    For each node~$i$ a time window $[s_i, t_i]$ is given that represents the earliest and latest time of
    arrival at customer $i$. 
    A demand $d_i > 0$ is associated with each customer $i$ and
    the total demand of all customers in one route should not exceed a maximum capacity $K$.
    The goal is to minimize the total cost of the routes while ensuring feasibility.

    Exact approaches in the literature are mostly based on the branch-price-and-cut
    paradigm~\cite{costaExactBranchPriceandCutAlgorithms2019}.
    Current state-of-the art algorithms are able to solve most instances of up to $200$ customers to proven optimality~\cite{pecinNewEnhancementsExact2017}.
    In practice, however, the size of the instances can be much larger and optimality guarantees are often not required.
    Therefore, there has been much interest in efficient heuristic and meta-heuristic algorithms that can
    find high-quality solutions within a reasonable amount of time.
    For VRPTW, meta-heuristics have proved highly effective; finding optimal or near-optimal solutions within seconds
    and can scale to problems with thousands of nodes.
    In the $12^{th}$ DIMACS challenge~\cite{dimacs2021cvrp}, the \emph{Hybrid Genetic Search}
    (HGS)~\cite{vidalHybridGeneticAlgorithm2012} meta-heuristic in an implementation by Kool et al.~\cite{koolHybridGeneticSearch}
    was the top performing submission with respect to primal solution quality.

    In the recent \textit{EURO meets NeurIPS Vehicle Routing Competition}~\cite{euromeetsneurips2022}, a generalization
    of VRPTW was posed.
    The \emph{Dynamic VRPTW} (DVRPTW) is defined over epochs $t \in {1, 2, \ldots, T}$; with each epoch $t$
    a set of customers is given.
    The task is to select \emph{directly after each epoch}, i.e., without knowledge on future epochs, a subset of customers
    to dispatch and a feasible route for these customers.
    All customers appearing in any epoch must be dispatched eventually, so when a customer is not chosen in some epoch
    $t'$, then it will reappear in all future epochs~$t > t'$ until it is dispatched.

    Additionally, due to time window constraints, some nodes need to be dispatched in the current epoch.
    These are marked as ``\emph{must-go}''.
    In particular, all customers appearing in the last epoch must be dispatched.
    The objective is to minimize the total drive times of all routes computed in all epochs.
    The original VRPTW can be seen as a special case of DVRPTW with only one epoch.

    In this work, we introduce a general method for DVRPTW based on the state-of-the-art meta-heuristic HGS.
    First, let us note that any method for VRPTW can be lifted
    trivially to the dynamic setting if we specify a strategy that
    chooses a fixed subset of customers to be dispatched at each
    epoch: simply perform, at each epoch, the VRPTW method on the
    fixed subset of customers in order to find high-quality routes.
    In our experiments we include several baselines that follow precisely this scheme, calling HGS at each epoch on a fixed set of customers.
    
    By contrast, the motivation for our work is to extend HGS's solution management and local search techniques in order to decide \emph{in an integrated fashion}
    which customers to dispatch and which routes to use for serving the chosen customers.
    The resulting algorithm DHGS is called at each epoch and does not require any pre-training on existing data.

    The paper is organized as follows. \autoref{sec:hgs} presents a succinct description of the HGS algorithm.
    Our adaptions to DVRPTW are described in \autoref{sec:adapting}.
    In \autoref{sec:comp} we investigate how these adaptations perform computationally by comparing them to different baselines, and give concluding remarks.


    \section{Hybrid Genetic Search}\label{sec:hgs}

    Hybrid Genetic Search was first introduced by Vidal et
    al.~\cite{vidalHybridGeneticAlgorithm2012} for
    vehicle routing and has proven to be effective on many variants of
    the problem~\cite{vidalUnifiedSolutionFramework2014}.
    As a genetic algorithm, it aims to balance solution quality and
    diversity and uses the corresponding metaphors of referring to
    solutions as \textit{individuals}, the set of solutions as a
    \textit{population}, and \textit{fitness} to refer to a measure of
    solution's quality.

    The algorithm maintains two separate sets for feasible and infeasible solutions.
    In the beginning, an \textit{initial population} is created using multiple construction heuristics.
    At each iteration, two individuals are chosen based on their fitness and a \textit{crossover operator} is
    applied to them in order to create a new individual.
    This individual is then improved using \textit{local search} operators and added back to the corresponding population.

    The controlled evaluation of infeasible solutions is one of HGS's important ingredients, as optimal solutions are likely
    to exist near the border of feasibility.
    This evaluation is performed through penalizing the violation of constraints.
    The (penalized) cost of a solution is composed of two parts: one for the total cost of the routes,
    the other for infeasibility of the routes.
    This is then combined with a measure of diversity to represent the overall fitness of the solution.

    The overall HGS algorithm is presented in \autoref{alg:hgstw}.
    For the purposes of a compact presentation, many of the details of the algorithm are not mentioned here. For a detailed description we refer to \cite{vidalHybridGeneticAlgorithm2012}.
    The parts relevant to our work are detailed in the next section along with our adaptation to DVRPTW.

    \vspace*{-1.5em}

    \begin{algorithm}[th]
        initialize population with random solutions improved by local search\;
        \While{number of iterations without improvement and time within limits}{
            select parent solutions $P_1$ and $P_2$\;
            apply all crossover operators on $P_1$ and $P_2$ and choose offspring $C$ with minimum penalized cost\;
            improve offspring $C$ using local search\;
            insert $C$ into the respective subpopulation based on feasiblity\;
            \If{C is infeasible}{
                with $50\%$ probability, repair $C$ (local search) and insert it into respective subpopulation\;
            }
            \If{maximum subpopulation size reached}{
                select survivors\;
            }
            adjust penalty coefficients for infeasibility\;
        }
        \Return best feasible solution\;
        \caption{The HGS Algorithm (adapted from~\cite{vidalHybridGeneticAlgorithm2012}).}\label{alg:hgstw}
    \end{algorithm}

    \vspace*{-3em}

    \section{Adapting HGS for the Dynamic VRPTW}\label{sec:adapting}

    In this section we present our \emph{Dynamic HGS} (DHGS) algorithm to be called at each epoch of a DVRPTW instance.
    The starting point of the algorithm is the HGS implementation by Kool et
    al.~\cite{koolHybridGeneticSearch} for VRPTW, which in turn relies
    on the open-source implementation of HGS for capacitated vehicle
    routing by Vidal~\cite{vidalHybridGeneticSearch2022}.
    In the following, we discuss specific parts of HGS along with their adaptation to DHGS.

    \emph{Solution Representation.} In HGS, solutions are represented using the so-called \emph{giant-tour}
    representation.
    Routes in a solution are concatenated without specifying the visits to depots.
    From this representation, optimal positions for the depots can be computed using a linear time
    SPLIT algorithm~\cite{vidalTechnicalNoteSplit2016}.
    While for VRPTW all giant tours must consist of the same number of nodes, a solution for one epoch of DVRPTW can include only a subset of the nodes.
    Hence, for the dynamic setting, we allow giant tours to have variable size.



    \emph{Initial Population.} At the outset, HGS creates an initial set of solutions
    via multiple construction heuristics.
    Most of the initial solutions are generated using the \textit{random} strategy;
    A random ordering of the customers is generated, then split into routes using~\cite{vidalTechnicalNoteSplit2016}.
    Local search is then applied and the individual is added to the corresponding population.
    In DHGS, we construct the initial population in a similar manner by generating random orderings of the customers.
    Subsequently, we remove a subset of the optional customers, each with fixed probability.
    In our implementation we use probabilities $p \in \{0, 0.1, 0.5, 1 \}$ to create (at most) four solutions from each random ordering.

    \emph{Crossover.} This operation of combining two (parent) solutions into a new (offspring) solution has
    the goal of exploring a new subspace of solutions.
    The new solution should be sufficiently different from the parent solutions while maintaining reasonable quality.
    HGS uses the \emph{Ordered Crossover} (OX)~\cite{oliverStudyPermutationCrossover1987} operator,
    which copies a random contiguous fragment of the giant-tour representation of the first parent solution and
    then completes it
    with the missing nodes in the order in which they appear in the second parent.
    In DHGS we apply the same technique on solutions of variable size.
    Note that this never lead to solutions of smaller size.
    While this could in theory affect diversity negatively, this effect
    is hopefully mitigated by the \emph{delete} local search operator explained below.

    \emph{Fitness.} In HGS, the fitness of a solution (individual) is a score that reflects the cost and (in)feasibility of a solution.
    The difficulty in the dynamic setting is to find a good proxy for the quality of routes on the chosen set of customers \emph{plus}
    the effect of this choice on future epochs.

    First, in order to avoid that solutions with fewer nodes appear to have better cost, we normalize the cost with respect to
    the number of nodes chosen $n$.
    Second, to reflect the effect on future epochs, we add a \emph{lateness} measure $S_{late}$ calculated as the sum of end of the
    time windows for all dispatched customers; this favor customers that are more likely to become must-go
    nodes in the near future.
    Third, we modify the normalization by dividing by a number representing the progress with respect to the
    total number of epochs
    in order to incentivize the algorithm to dispatch more nodes as it approaches the last epoch.
    To summarize we calculate fitness as
    \[
        \frac{p_{dist} \cdot S_{dist} + p_{cap} \cdot S_{cap} + p_{time} \cdot S_{time} + p_{late} \cdot
        S_{late}}{(\frac{t-1}{T} + 1) \cdot (n + 1)},
    \]
    where $S_{dist}$ is the total distance of the routes, $t$ is the index of the current epoch, $T$ is the total number of epochs, $S_{cap}$ and $S_{time}$ are the penalized infeasibility
    measures computed as in HGS for capacity and time window constraints, respectively.
    The non-negative weights.
    In our implementation $p_{late}$ is set to $100$ and $p_{dist}$, $p_{cap}$, and $p_{time}$ are managed by HGS as described
    in~\cite{vidalHybridGeneticAlgorithm2012};
    they are updated to balance the search between increasing feasibility of certain violated constraints
    and improving cost.

    \emph{Local Search.}
    Local search heuristics are an essential ingredient of HGS to improve solutions found from the initial population and crossover.
    They work by applying local improvement operators until no improvement can be achieved.
    These operators range from simple relocate-and-swap steps to more complex modifications.
    HGS includes multiple local search heuristics, that are applied in the order of increasing computational complexity.
    We add three new operators to
    allow for manipulating the set of dispatched nodes in a solution:
    \begin{itemize}
        \item \texttt{delete} removes an optional customer if this improves feasibility.
        \item \texttt{add} inserts a customer if this does not affect the cost significantly.
        \item \texttt{swap-out} exchanges a customer that is part of the solution with one that is not used if it
        improves the original HGS' score $S_{dist} + S_{cap} + S_{time}$.
    \end{itemize}
    We run the new heuristics first to modify the set of dispatched customers; subsequently, we improve the routes using the
    original HGS heuristics.


    \vspace*{-0.5em}
    \section{Computational Results}\label{sec:comp}

    Our DHGS implementation is based on the HGS solver for VRPTW provided for the
    EURO meets NeurIPS Vehicle Routing Competition~\cite{euromeetsneurips2022}.
    We evaluate the algorithm on the full set of 250~instances released in the competition against the following
    baselines provided by the competition organizers:
    \begin{itemize}
        \item \texttt{greedy} dispatches all customers in each epoch.
        \item \texttt{lazy} dispatches only the must-go customers.
        \item \texttt{random} dispatches the set of must-go customers in addition to some optional customers
        chosen at random.
        \item \texttt{oracle} has access to all future information and solves the problem in hindsight as one large VRPTW\@.
        The oracle solution is not guaranteed to be a global optimum since it uses the HGS heuristic; still, it can be viewed as an approximate lower bound for the best solution quality that can be achieved.
        \item \texttt{supervised} is an imitation learning approach based on graph neural networks trained on the oracle
        solver's decisions.
        \item \texttt{dqn} is a deep-Q neural network trained using reinforcement learning.
    \end{itemize}
    These baselines all use HGS as a subroutine and differ only in the strategy to choose the subset of customers to
    dispatch in each epoch.

    Experiments were conducted on a cluster of identical machines equipped with Intel(R) Xeon(R) Gold 5122
    processors with 3.6GHz and 96GB of RAM; each chip has 4~cores, but experiments were run in single-threaded mode.
    The time limit for each solver is $5$ minutes per epoch.
    The \texttt{oracle} strategy runs the \texttt{greedy} strategy first to
    get a good initial solution then is given $10$ minutes to run HGS on all nodes from all epochs.
    All solvers are initialized with the trivially feasible singleton solution, i.e., a collection of routes
    visiting only a single node and returning to the depot.

\begin{figure}[th]
        \centering

        \includegraphics[scale=0.45,trim=0cm 1cm 0cm 1cm, clip=true]{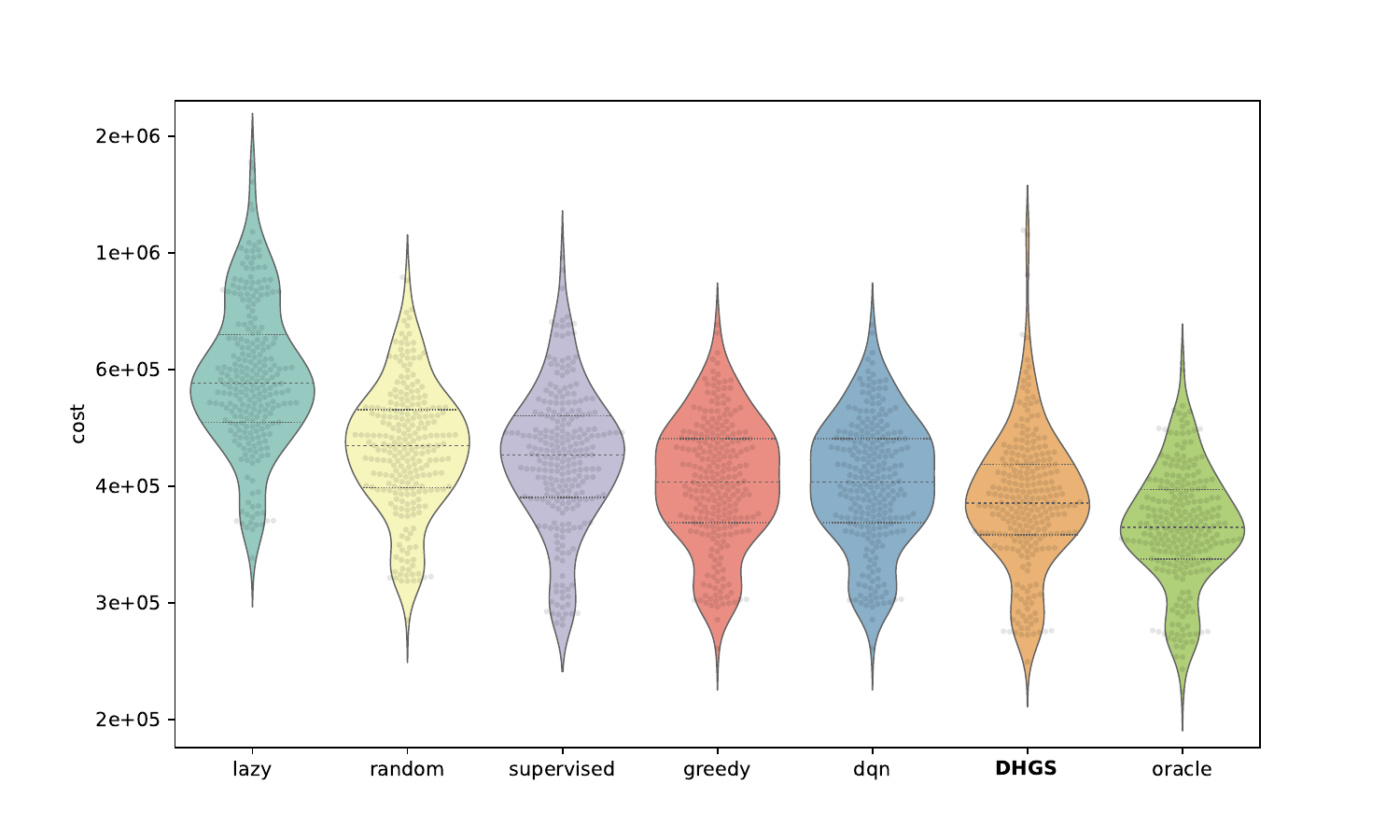}
        \caption{A violin plot comparing the total cost objective of the different solvers over all
        250~instances. Each grey point represents an instance and its total cost for the corresponding solver. For
        each solver, the violin plot shows the distribution of the points, with horizontal lines marking the median
        and the $1^{st}$ and $3^{rd}$ quartiles.}
        \label{fig:box}

    \end{figure}

    The results are summarized in \autoref{fig:box}.
    On average, DHGS improves the cost objective by 7.82\% when the
    median cost is compared to that of the top
    performing baselines \texttt{greedy} and \texttt{dqn}.
    The gap to the \texttt{oracle} strategy is 10.37\%.
    In addition, the raw performance data also shows that the size of the instance does not affect the relative ranking of the methods.

    While DHGS outperforms all other strategies on average and in most cases, its performance shows higher variance.
    On $7$ out of the $250$ instances, DHGS (and only DHGS) achieves better quality than the oracle
    baseline.
    However, on $19$ instances DHGS is also worse than \texttt{greedy} and
    \texttt{dqn}, and on $3$ instances it has the highest cost of all strategies.
    In these $3$ cases, there is an epoch where DHGS gets stuck at the trivial initial solution, which uses a separate
    vehicle for each customer.

    \vspace*{-2em}

%


    \vspace*{-1.5em}

    To conclude, the results demonstrate that the adapted DHGS algorithm can perform favorably against different baselines, providing an effective
    general method for DVRPTW that does not require any pre-training.
    However, the dependency of DVRPTW on online data also makes it a natural test case for the application of machine learning.
    The results show that this is not straightforward.
    The \texttt{supervised} baseline, although trained on the \texttt{oracle}
    results, performs worse than the trivial \texttt{greedy} strategy.
    The reinforcement learning baseline \texttt{dqn} effectively learned the \texttt{greedy} strategy.

    However, many of the top performing solvers in the competition were based on machine learning
    and demonstrate that more sophisticated approaches can successfully harness the potential of machine learning~\cite{kleopatra,team_sb,milestogobeforewesleep}.
    Some follow the simple scheme of applying HGS as a subroutine after performing ML-based customer selection.
    Hence, one interesting question for future research is if combining these ideas with DHGS can
    improve performance and robustness further.




    \bibliographystyle{splncs04}
    \bibliography{main}

\end{document}